\documentclass[conference]{IEEEtran}
\IEEEoverridecommandlockouts
\usepackage{cite}
\usepackage{amsmath,amssymb,amsfonts}
\usepackage{algorithmic}
\usepackage{graphicx}
\usepackage{textcomp}
\usepackage{xcolor}
\def\BibTeX{{\rm B\kern-.05em{\sc i\kern-.025em b}\kern-.08em
    T\kern-.1667em\lower.7ex\hbox{E}\kern-.125emX}}
\begin{document}

\title{5G Traffic Prediction with Time Series Analysis\\
}




\author{
Nikhil Nayak\textsuperscript{1*}\thanks{*Correspondence to: Nikhil Nayak \texttt{<nayak.nikhil2608@gmail.com>}}, 
Rujula Singh R\textsuperscript{1}, 
Rameshwar Garg\textsuperscript{1}, 
Varun Danda\textsuperscript{1}, 
Chandana Kiran\textsuperscript{1}, 
Kaustuv Saha\textsuperscript{2} \\~\\
\textsuperscript{1}R.V. College of Engineering\quad\textsuperscript{2}Nokia
}

\maketitle

\begin{abstract}
In today’s day and age, a mobile phone has become a basic requirement
needed for anyone to thrive. With the cellular traffic demand increasing so
dramatically, it is now necessary to accurately predict the user traffic in cellular
networks, so as to improve the performance in terms of resource allocation and
utilisation. Since traffic learning and prediction is a classical and appealing field,
which still yields many meaningful results, there has been an increasing interest in
leveraging Machine Learning tools to analyse the total traffic served in a given region,
to optimise the operation of the network. With the help of this project, we seek to
exploit the traffic history by using it to predict the nature and occurrence of future
traffic. Furthermore, we classify the traffic into particular application types, to increase
our understanding of the nature of the traffic.
By leveraging the power of machine learning and identifying its usefulness in
the field of cellular networks we try to achieve three main objectives - classification of
the application generating the traffic, prediction of packet arrival intensity and burst
occurrence. The design of the prediction and classification system is done using Long
Short Term Memory (LSTM) model. The LSTM predictor developed in this experiment
would return the number of uplink packets and also estimate the probability of burst
occurrence in the specified future time interval. For the purpose of classification, the
regression layer in our LSTM prediction model is replaced by a softmax classifier
which is used to classify the application generating the cellular traffic into one of the
four applications including surfing, video calling, voice calling, and video streaming.
\end{abstract}


\section{INTRODUCTION}
In the current era, mobile phones have become a basic requirement needed for everyone to carry, on a day-to-day basis. With the cellular traffic demand increasing so dramatically, it is now necessary to accurately predict the user traffic in cellular networks, so as to improve the performance in terms of resource allocation and resource utilisation. Since traffic learning and prediction is a classical and appealing field, which still yields many meaningful results, there has been an increasing interest in leveraging Machine Learning to analyse the total traffic served in a given region, to optimise the operation of the network. Based on the analysis, we can exploit the traffic history by using it to predict the nature and occurrence of future traffic. Furthermore, we classify the traffic into particular application types, to increase our understanding of the nature of the traffic. 

\section{OBJECTIVE}

With the help of Machine Learning, we can identity various trends and patterns in 5G cellular networks. In this project we strive to achieve four main objectives as follows:
\begin{itemize}
\item Prediction of burst occurrence, i.e., prediction of whether or not there will be a sudden spike in traffic flow at a given moment of time.
\item Prediction of the intensity of packet arrival at a given instant of time.
\item Analysis of inter-arrival time for all packets within a total period of 7 days.
\item Categorise the packet flow into 4 parts, using a clustering algorithm.
\end{itemize}

\section{RELATED WORK}
In [2], the dynamics of cellular traffic along with the nuances for predicting future demands to improve resource efficiency are understood. Various methods of time series forecasting for cellular traffic using Convolutional and Recurrent Neural Networks are explored in [6]. In [8], the spatial and temporal correlations of the cellular traffic in different time periods and neighbouring cells, respectively, have been explored using neural networks in order to improve the accuracy of traffic prediction. The authors of [7] propose a method to combine CNNs and RNNs to extract more information and yield better results in the field of time series forecasting for cellular traffic. These prior studies collectively highlight the growing importance of deep learning approaches for capturing complex temporal dependencies in network traffic. However, many of them overlook practical preprocessing steps and feature engineering strategies that are critical for achieving real-world performance. In [4, 9], solutions for time series analysis using LSTMs are presented, but the analysis is weak and they are lacking in technical details and do not mention the set of features that have been used for the training and analysis. Various other papers [12, 10, 11, 4, 9, 5] have been used for technical details and ideas, helping us with the work we have done as a part of this project. The paper [3] serves as our base and we use it to model our implementation.

\section{PRELIMINARIES}

\subsection{5G Data}\label{AA}
5G is the fifth generation technology standard for broadband cellular networks. It is designed to increase speed, reduce latency, and improve flexibility of wireless services. 5G is a unified, more capable air interface. It has been designed with an extended capacity to enable next-generation user experiences, empower new deployment models and deliver new services. It’s goal is to connect virtually everyone and everything together including machines, objects, and devices.

\subsection{Time Series Analysis }
A time series is simply a series of data points ordered in time. In a time series, time is often the independent variable and the goal is usually to make a forecast for the future. Time Series analysis can be useful to see how a given asset, security or economic variable changes over time. Using characteristics of time series like autocorrelation, stationarity and seasonality, predictions for future time intervals are made based on the data at previous instances of time. Meaningful statistics and characteristics of data can also be brought out using time series analysis.

\subsection{Time Series Analysis For Cellular Traffic}
Time Series Analysis can be used to detect malicious data in cellular traffic and detect any anomalies. It can be used to predict and model the behaviour of cellular networks, so as to understand the traffic and cause of traffic. This technique is furthermore used for dynamic resource allocation by predicting unusual network load and then using other models to reallocate traffic accordingly. Time Series analysis is best suited to predict occurrence of bursty traffic and to estimate the intensity of the burstiness.

\subsection{Long Short Term Memory}
Long Short-Term Memory (LSTM) networks are a type of recurrent neural network capable of learning order dependence in sequence prediction problems. LSTMs are capable of learning and using long term dependencies. LSTM units have a cell, an input gate, an output gate and a forget gate. The cells remember information over arbitrary time intervals and the three gates regulate the flow of information into an out of the cell. They can be used for text analysis, speech recognition, language modulation, time series analysis and many other applications.

\subsection{LSTM - Time Series Analysis}
LSTMs can be used for time series analysis where they look at time series data and learn to make predictions from them. The LSTMs take in multiple rows of data as a subset and then learn from that in every step. These variants of RNNs have proven to be better learners and capture time related trends much better than most of its counterparts.

\section{METHODOLOGY }
In this section, we focus on the explanation of various background requirements, the proposed approach and the implementation of the final model.

\subsection{Data Preprocessing \& Cleaning}
The dataset has been taken from [1]. Here, raw data was generated using a variety of applications and captured using WireShark. This data was stored in MATLAB Class objects and had to be extracted using MATLAB. The extracted data had to be converted to the right format for columns like Timestamp and NaN values had to be dropped. The source and destination IP Addresses were analysed to identify the user addresses and the tower address. Based on the IP addresses and direction of flow, the packets were labelled as Uplink and Downlink Packets. The packets are grouped together based on the timestamp to incorporate all the packets within a given period of time, for example, 1s. The total number of uplink and downlink packets are aggregated for every time group and the length of packets are also taken into account for further processing. The total amount of data transmitted in that given time period (ex. 1s) is then calculated. The final features are Uplink/Downlink, Count, Size of Uplink/Downlink, Uplink-Downlink Ratio, Protocol and the total amount of data transmitted serves as the target variable.

\subsection{Network Architecture}
We use a Deep Learning model to predict whether the traffic will have sudden spikes in intensity and also how severe these particular spikes are. For this, we need a thorough network of neurons. We employ the following as a part of our model.
\begin{itemize}
 \item An LSTM layer comprising of 100 units/elements.
 \item The LSTM layer is followed by a Fully Connected layer.
 \item The output layer is a Regression layer as we need to predict real values.
\end{itemize}

\section{PROPOSED APPROACH}
\begin{itemize}
    \item The cellular traffic dataset contains the incoming/outgoing packets captured at the Internet protocol (IP) level with help of packet capture tool e.g. WireShark.
\item Prediction of the burst occurrence is done using Long Short Term Memory (LSTM) model.
\item 	Root Mean Square Error (RMSE) and a mapping of the actual values to the predicted values are chosen as the performance indicators for the prediction of traffic intensity.
\item Accuracy, recall, F1 Score and precision are the performance indicators for the prediction of the occurrence of a burst.
\item 	The LSTM network returns the number of uplink packets and estimates the probability of burst occurrence in a specified time interval. Occurrence of a burst is predicted by setting a burst threshold.

\end{itemize}

\begin{figure}[http]
\centerline{\includegraphics[width=8cm,height=5cm]{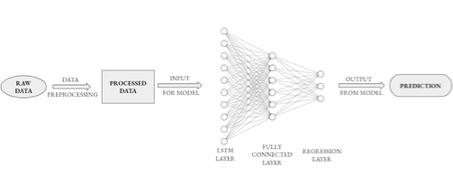}}
\caption{A Flowchart Depicting Our Proposed Approach}
\label{fig1}
\end{figure}

A brief overview of the entire process can be seen in Fig.1. It is a Flowchart which brings out the significance of various steps that we have taken in this project.

\section{IMPLEMENTATION}
\subsection{Preprocessing Implementation}

\begin{itemize}

\item FeatureExtraction:
This is to convert the raw data into meaningful data using Feature Extraction techniques.
\item ZeroPadding:
The is used to pad the data with 0 rows, so as to make the data meaningful without any inconsistencies.
\item ProPadding:
The is used to pad zeroes in a special way, so as to reduce the bias towards 0s and prevent the meaningful data from being treated like outliers.
\end{itemize}

\subsection{Implementation of LSTM}
\begin{itemize}
\item LSTM Implementation 1x60:
This is used to train and evaluate the LSTM on data with bin size of 1s and history of 60 rows.
\item
LSTM Implementation 3x20:
This is used to train and evaluate the LSTM on data with bin size of 3s and history of 20 rows.
\item
LSTM Implementation 60x1:
This is used to train and evaluate the LSTM on data with bin size of 60s and history of a single row.
\item
ProPaddingImplementation:
This is used to train the LSTM on data padded with limited 0s to reduce bias, with bin size of 1s and history of 10 rows.

\end{itemize}

\subsection {Implementation of LSTM \& Clustering}
\begin{itemize}
\item
LSTM Implementation NonZeroPrediction Burstiness:
This is used to predict only nonzero values, so as to capture the trend of the data and not get influenced by the large number of zeroes, using bin size of 1s and history of 10s.
\item
KMeans CLutering Cellular Traffic Variation:
This is used to compare the results of KMeans clustering with and without UplinkDownlink Ratio for bin sizes of 1s, 30s and 60s with a history of a single row.
\end{itemize}

\section{RESULTS AND DISCUSSION }
\subsection{Evaluation Metrics}
To test the goodness of the model, two evaluation metrics were used. The evaluation metrics encompass both the mathematical aspects as well as visual aspects:
Root Mean Squared Error (RMSE),
it is the standard deviation of the residuals. It tells you how concentrated the data is around the line of best fit. Root mean square error is commonly used in climatology, forecasting, and regression analysis to verify experimental results.
Graph,
the graph comparing the predicted values for Uplink Count with the Actual values of the Uplink Count is a very good way to evaluate whether or not the predictions are accurate and are fitting the general trend line.
RMSE does not prove to be good metric by itself. While predicting values, if the model is biased to zeroes and flatlines by predicting only zero values, then the error might not be very large, as there are very few peaks and do not contribute to errors significantly. On the other hand, even if the RMSE is very high, we cannot say that the model has failed, because the large error might be the cumulative result of extremely small and acceptable deviations from the zero value, for every single zero value. We try to gauge our model with the help of a graph, to see the goodness of fit and then decide the next step of action. Even the graphical method cannot be solely sufficient as we cannot estimate the true error by looking at just the graph. Hence, we use the coupled combination of the RMSE score and the graph, to get a true understanding of the model and its behaviour.

\subsection{Results of LSTMs}
\begin{itemize}
    \item LSTM  Implementation 1x60:
The model gave a very good error score but the graph was a flat line and the model had learnt to predict only zeroes, as a result of which, the RMSE was quite low. This shows that the model was being highly influenced by the zero padded rows and had developed bias.
\item LSTM  Implementation 3x20:
This model also faced the same issue as the above model, where it developed a bias towards the zeroes and predicted a flat line with low error score. This led to the understanding that a larger bin size might be helpful.
\item LSTM  Implementation 60x1:
The model successfully learnt the trend line, but could not predict bursts and could not reach the peaks as it was still being trained on too many zeroes. The error was quite high as it was not predicting zeroes for the zero values and was not reaching peaks as well. It was stuck in the middle. This showed the total 60s interval was not a suitable parameter.
\item ProPaddingImplementation:
To eliminate the bias introduced by the zero padded rows, the padding was done in a special way. The model gave good results for RMSE but the trend line was still flat, as the bin size was 1s and the model was learning to predict zeroes as most bins were empty due to the small size of bins.

\end{itemize}

\subsection{Results of LSTM and K-Means}
\begin{itemize}
    \item LSTM  Implementation NonZeroPrediction Burstiness:
Since the previous models were being trained on zero values, we adjusted the training set so that the model learns from the zero padded history but does not get influenced by trying to predict the zeroes. The model gave excellent results as the RMSE had been significantly lowered and even the trend line was matched almost completely. The burstiness was also predicted by applying a threshold to the predicted value. The threshold was taken as the mean of all the Uplink packet counts in the training set. The accuracy of the prediction of burst was also measured and this gave us excellent results. This shows that our model was successfully predicting bursts, as the classification metrics suggested, and was able to predict the intensity of the traffic as well, as is indicated by the error score and the graph. The confusion matrix can be calculated using the values of the parameters in the classification report. The results of the model can be seen in Fig.2.

\begin{figure}[http]
\centerline
{\includegraphics[width=9cm,height=8cm]{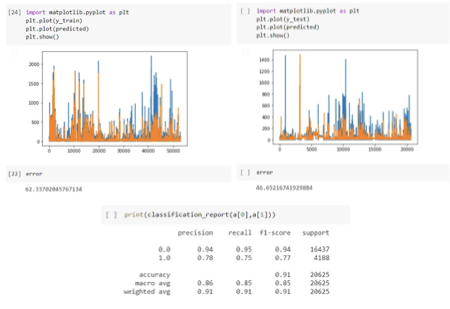}}
\caption{The RMSE and the Graph of the trend line for the Training and Testing sets respectively, followed by the Classification Report, quantifying the quality of the burst prediction.}
\label{fig2}
\end{figure}

\begin{figure}[http]
\centerline
{\includegraphics[width=7cm,height=5cm]{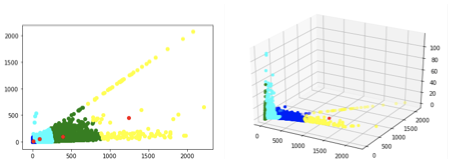}}
\caption{The clusters and their centres for a time bin of 1 second.}
\label{fig3}
\end{figure}

\begin{figure}[http]
\centerline
{\includegraphics[width=7cm,height=5cm]{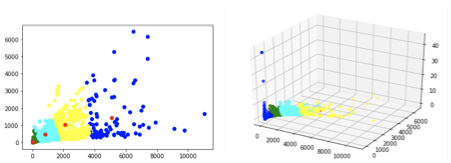}}
\caption{The clusters and their centres for a time bin of 30 seconds}
\label{fig4}
\end{figure}

\item KMeans CLutering Cellular Traffic Variation:
Clustering was performed on the entire dataset for various time bins employing various feature sets. The zero padded values were not included as they do not provide any value to the clustering algorithm. Time bins of 1s, 30s and 60s were used, each with two different subsets of features. The first subset comprises of the ‘UplinkCount’ and the ‘DownlinkCount’. The second subset adds an additional feature called the ‘Uplink/DownlinkRatio’. The general trend of clustering is similar across all the variations. Although the clustering could not be validated because of the lack of target variables, it is quite clear that the clustering was successful as the clusters in all the cases were quite similar and followed the same general trend. Fig.3 focuses on clustering the data when it was divided into time bins of 1s. Fig.4 is used to portray the results of clustering obtained when the data was compacted into time bins of 30s and finally Fig.5 shows the results of clustering when the data was aggregated into bin sizes of a minute.

\begin{figure}[http]
\centerline
{\includegraphics[width=7cm,height=5cm]{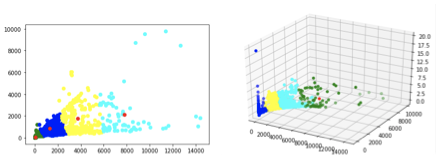}}
\caption{The clusters and their centres for a time bin of 60 seconds.}
\label{fig5}
\end{figure}

\end{itemize}

\section{CONCLUSION}
The prediction of traffic burst occurrence and intensity for cellular networks has been investigated, which leverages various machine learning tools. A comprehensive comparative analysis using prediction tools, such as LSTMs, has been carried out, under different traffic circumstances and design parameter selections. The practicality of the developed LSTM model for prediction of burstiness has also been investigated. Our project dives deep into various parameters to understand which model is the most suitable. Reasons for the success or failure of all the models have been understood and presented as a part of this project. Furthermore, data has been clustered into 4 applications, based on how the data was generated. The results are quite satisfactory and consistent with our theoretical observations.

\section{FUTURE WORK}
In the future, we would like to generate our own data as it would be more consistent and complete. We would also have the necessary information regarding the class of applications generating this data. This would help us in achieving better overall quantitative results, which includes, but is not limited to higher accuracy prediction for bigger peaks. Furthermore, supervised classification of the data, into the categories of applications generating the same can be done using our generated data.

Future prospects for this project can also include analysis of 5G data from various IOT devices, which we have currently omitted for the purpose of simplicity. Time Series Analysis oriented towards IOT can be carried out and feasible solutions to the various inherent problems present in the same field can be found.

\end{document}